\documentclass[11pt,a4paper]{article}
\usepackage{graphicx}
\usepackage[affil-it]{authblk}
\usepackage{grffile}
\usepackage{cite}
\usepackage{amsmath}
\usepackage{amssymb}
\usepackage{multirow}
\usepackage{booktabs}

\usepackage[ruled,vlined]{algorithm2e}

\newcommand{\argmin}{\arg\!\min}

\begin{document}
\date{}
\title{Multi-Objective Design of State Feedback Controllers Using Reinforced Quantum-Behaved Particle Swarm Optimization}
\author{Kaveh Hassani%
\thanks{email: \texttt{kaveh.hassani@uottawa.ca}; Corresponding author}  and Won-Sook Lee%
\thanks{email: \texttt{wslee@uottawa.ca}\\
The complete paper appears in Applied Soft Computing\\
DOI: http://dx.doi.org/10.1016/j.asoc.2015.12.024 }}
  
\affil{School of Computer Science and Electrical Engineering\\ University of Ottawa, Canada}
\maketitle


\begin{abstract}
\noindent
In this paper, a novel and generic multi-objective design paradigm is proposed which utilizes quantum-behaved PSO(QPSO) for deciding the optimal configuration of the LQR controller for a given problem considering a set of competing objectives. There are three main contributions introduced in this paper as follows. (1) The standard QPSO algorithm is reinforced with an informed initialization scheme based on the simulated annealing algorithm and Gaussian neighborhood selection mechanism. (2) It is also augmented with a local search strategy which integrates the advantages of memetic algorithm into conventional QPSO. (3) An aggregated dynamic weighting criterion is introduced that dynamically combines the soft and hard constraints with control objectives to provide the designer with a set of Pareto optimal solutions and lets her to decide the target solution based on practical preferences. The proposed method is compared against a gradient-based method, seven meta-heuristics, and the trial-and-error method on two control benchmarks using sensitivity analysis and full factorial parameter selection and the results are validated using one-tailed T-test. The experimental results suggest that the proposed method outperforms opponent methods in terms of controller effort, measures associated with transient response and criteria related to steady-state.
\end{abstract}

\section{Introduction}
Optimal control theory refers to the controller design patterns that simultaneously satisfy the physical constraints of the controlled process and optimize some predetermined performance criteria. The evolution of optimal control theory has led to the emergence of linear quadratic regulators (LQR) -- an optimal multivariable feedback control approach that improves the stability and minimizes the excursion in state trajectories of a system while requiring minimum controller effort. Applying LQR technique to a controllable linear time-invariant (LTI) system results in a set of optimal feedback gains that minimizes a quadratic criterion and stabilizes the system \cite{1-lewis1995optimal}. The LQR approach has been utilized in vast variety of real world engineering applications such as but not limited to missile guidance \cite{2-savkin2003problem,3-savkin2003problem}, flight control \cite{4-khan2011optimized}, multiple spacecraft formation \cite{5-starin2001design}, controlling unmanned vehicles \cite{6-hernandez2011non}, active car suspension \cite{7-sam2000lqr}, ABS break system \cite{8-johansen2003gain}, power converters \cite{9-poodeh2007optimized,10-olalla2009robust}, active power filter \cite{11-kedjar2009dsp}, and tuning PID controllers \cite{12-kumar2013lqr}. Essentially, LQR controllers minimize a quadratic cost function also known as performance index that consists of two penalty matrices including state (Q) and control (R) weighting matrices. These two parameters are main design parameters to be selected by the designer and greatly influence the behavior of the controller. It is worth noting that it is not a trivial task to decide these two matrices. In general, deciding the LQR parameters for a given process is a continuous, multimodal and multi-objective optimization problem. For the simplicity purposes, in most of the applications the problem is modeled as a single-objective which limits the designer to only one configuration. In the proposed approach, the problem is considered as a multi-objective problem and as a result the designer is provided with a set of Pareto optimal solutions which lets her to decide the target solution based on some practical preferences.

Traditionally, weighting matrices are determined by the trial-and-error method in which a domain expert adjusts the weighting matrices intuitively and then refines them iteratively to obtain a satisfactory performance. This method is not feasible for high-dimensional systems and even for simpler systems is labor-intensive and time-consuming. Bryson's method \cite{13-johnson1987recent} is another iterative method in which initial state and feedback variables are normalized with respect to their largest permissible and then are utilized to initialize the weighting matrices. Then, similar to trial-and-error method, the weighting matrices are gradually refined to approach the minimum index value. Pole placement \cite{14-saif1989optimal} is another popular technique for determining the weighting matrices in which the matrices are decided based on the given poles. This approach neither guarantees a good performance nor the satisfaction of the constraints. Other approaches have been proposed as well such as utilizing asymptotic modal properties \cite{15-harvey1978quadratic} and expressing the system as an explicit function of the weighting matrix elements \cite{16-tyler1966use}. Yet they suffer from the similar deficiencies. These classic approaches are labor-intensive, time-consuming and do not guarantee the expected performance. They only aim to minimize the quadratic performance index and ignore the other competing or incommensurable control objectives such as minimizing the overshoot, rise-time, settling-time, and the steady-state error. 

In order to tackle these problems, some studies have utilized soft computing techniques including but not limited to particle swarm optimization (PSO) \cite{17-tsai2013variable}, artificial bee colony (ABC) \cite{18-changhao2013artificial}, ant colony optimization (ACO) \cite{19-douik2008optimised}, genetic algorithm (GA) \cite{20-wongsathan2009application}, differential evolution (DE) \cite{21-liouane2012probabilistic}, memetic algorithm (MA) \cite{22-zhang2011application}, artificial immune systems (AIS) \cite{23-ramaswamy2007optimal}, imperialist competitive algorithm (ICA) \cite{24-rakhshani2012intelligent}, neural networks \cite{25-wiklendt2009small}, and fuzzy systems \cite{26-tao2010design}. These methods can explore the search hyperspace in an informed manner and converge to the optimal solutions in a few iterations using a combination of knowledge sharing and individual explorations. The main problem with most of the computational intelligence techniques is that they are prone to premature convergence which causes them to get trapped within the local optima. 

In this paper, a novel and generic multi-objective design paradigm is proposed which utilizes a global convergence guaranteed variation of particle swarm optimization (PSO) called quantum-behaved PSO (QPSO) for deciding the optimal configuration of the LQR controller for a given problem considering a set of competing objectives including the quadratic performance index, overshoot, rise-time, settling-time, steady-state error, and integrated absolute error. The proposed method is called reinforced multi-objective quantum-behaved PSO (RMO-QPSO). The rationale behind the selection of QPSO as the core optimizer is as follows. (1) The experimental studies suggest that PSO as the predecessor of QPSO outperforms GA, DE, MA, ACO, and ABC in terms of success rate, solution quality and processing time \cite{27-civicioglu2013conceptual,28-elbeltagi2005comparison}. It is also experimentally shown that PSO is scalable, requires less computational resources, and its processing time grows at a linear rate with respect to the size of the problem \cite{29-akay2013study}. (2) it is theoretically guaranteed that the QPSO converges to the global optimum; (3) it is less sensitive to the bias problem as it has only one parameter, and (3) it outperforms other PSO variations in finding the optimal solutions \cite{30-sun2012quantum}.

There are three main contributions introduced in this paper as follows. (1) The standard QPSO algorithm is reinforced with an informed initialization based on the simulated annealing and Gaussian neighborhood selection mechanism. (2)  It is also augmented with a local search strategy which integrates the advantages of memetic algorithm into conventional QPSO. (3) Finally, an aggregated dynamic weighting criterion is introduced that dynamically combines the soft and hard constraints with control objectives to provide the designer with a set of Pareto optimal solutions and lets her to decide the target solution based on practical preferences. As far as the authors' knowledge is concerned, this is the first time that a multi-objective derivation of PSO is applied for deciding the configuration of LQR controllers.
 
Without loss of generality, RMO-QPSO is utilized to decide the configuration of LQR controllers for two control benchmarks including: (1) stabilizing an inverted pendulum system, and (2) controlling a flight landing system. In order to have comparative studies, nine different techniques (i.e. one trial-and-error based method, one gradient based technique, and seven stochastic meta-heuristics) including the trail-and-error method, Levenberg-Marquardt optimization (LM), GA, DE, ABC, PSO, QPSO, chaotic PSO (CPSO), and adaptive inertia weighted PSO (AIWPSO) are utilized to decide the LQR parameters of the same benchmarks. In order to mitigate the bias problem and have a fair comparison among the different techniques, full factorial parameter selection and sensitivity analysis \cite{31-lee2013sensitivity} are exploited for all of the applied techniques to find the best parameter setting including population size, iteration number, etc. The comparative results are also validated using one-tailed T-test to investigate whether the experimental results are statistically significant.
 
The paper is organized as follows. Section 2 provides the mathematical foundation of LQR controllers. In section 3 an overview of related works is presented. In section 4, we investigate the concept of multi-objective QPSO. In section 5, we present the proposed technique for optimal tuning of LQR controllers and in section 6 we discuss the experimental results. Section 7 concludes the paper.

\section{Linear Quadratic Regulators}
The LQR is an optimal multivariable feedback control approach that improves the stability and minimizes the excursion in the state trajectories of a system while requiring minimum controller effort. From a Mathematical point of view, for a controllable LTI system with a state-space model shown in \ref{eq:1}, the LQR approach constructs a linear state feedback law as depicted in \ref{eq:2}.

\begin{equation}
\label{eq:1}
\left\{ \begin{array}{rcl}
\dot{x}(t)=Ax(t)+Bu(t)\\
y(t)=Cx(t)+Du(t)
\end{array}\right.
\end{equation}

\begin{equation}
\label{eq:2}
u(t)=-Kx(t)
\end{equation}

In these equations, $x(t)$ denotes an n-dimensional state vector,$y(t)$ presents an r-dimensional output vector, and $u(t)$ is an m-dimensional control vector. $K\in{\Re^{n
\times m}}$ is the optimal state-feedback gain matrix. The control law in \ref{eq:2} minimizes the quadratic performance index shown in \ref{eq:3} which integrates the state and control energies through the time. In other words, it minimizes the distance between the process outputs and the desired outputs with minimum control energy.

\begin{equation}
\label{eq:3}
J=\int_{0}^{\infty}(x^TQx+u^TRu)dt
\end{equation}

In \ref{eq:3} $Q \in \Re{n \times n}$ denotes a symmetric positive semi-definite state weighting (state penalty) matrix and $R \in \Re{m \times m}$ denotes a symmetric positive definite control weighting (control penalty) matrix. The control gain matrix $K$ is given by \ref{eq:4}.

\begin{equation}
\label{eq:4}
K=R^{-1}B^TP
\end{equation}

where $P$ is a unique symmetric positive semi-definite solution to the algebraic Riccati equation shown in \ref{eq:5}.

\begin{equation}
\label{eq:5}
PA+A^TP+Q-PBR^{-1}B^TP=0
\end{equation}

\section{Related Works}
One of the most successful yet less applied techniques for optimal designing of LQR controllers is PSO. The superiority of PSO over GA in finding optimal weighting matrices has been experimentally shown in some studies \cite{32-ghoreishi2012optimal,33-zeng2012pso}. In \cite{34-nor2012optimal} PSO, GA and trial-and-error methods are utilized to adjust LQR weighting matrices which is applied to controlling an aircraft landing flare system. It is concluded that LQR design based on PSO is more efficient and robust compared to other methods. In \cite{35-vardhana2009robust} authors compared the performance of the ordinary LQR, the LQR with prescribed degree of stability (LQRPDS) and the PSO-based LQR in controlling distribution static compensator and showed that PSO-based design results in best performance under different operating conditions.

In \cite{36-solihin2010comparison} a method is proposed to determine the weighting matrices using PSO with pole region constraint for controlling a flexible-link manipulator. In \cite{34-nor2012optimal} it is suggested that PSO-based LQR produces better result compared to trial-and-error approach for the active suspension system. In \cite{24-rakhshani2012intelligent} authors applied ordinary LQR, PSO-based LQR, AWPSO-based LQR and ICA-based LQR for optimal load frequency control and concluded that AWPSO-based LQR outperforms other approaches in terms of settling-time and maximum overshoot. A PSO-based optimal real-time LQR controller for stabilizing an inverted pendulum system is proposed in \cite{37-guoping2010lqr}. In \cite{38-amini2013wavelet} wavelet-PSO is proposed for tuning LQR controllers and is applied to an optimal structural control. In \cite{39-karanki2010particle} it is concluded that contrary to trial-and-error approach, a PSO-based state feedback controller does not have sub-optimal performance in case of partial state feedback. Hassani et al. \cite{40-hassani2014optimal} proposed a quantum-behaved PSO algorithm for tuning a given LQR controller. They exploited a conventional weighted aggregation of control objectives which can only provide the designer with one solution. More studies on PSO-based LQR design can be found in \cite{4-khan2011optimized,17-tsai2013variable,41-solihin2009pso,42-xiong2010simulation}.

They are a few studies that exploit multi-objective design for tuning LQR controllers. In \cite{45-li2008design} a multi-objective evolutionary algorithm is presented to control a double inverted pendulum system. They concluded that multi-objective approach results in shorter adjusting time and smaller amplitude value deviating from steady-state in comparison with the pole assignment design approach. \cite{46-nekoui2012weighting} also applies a multi-objective evolutionary algorithm to stabilize a double inverted pendulum system. Their results suggest that multi-objective approach results in shorter adjusting time and smaller amplitude value deviating from steady-state in comparison to non-dominated sorting GA. As far as the authors' knowledge is concerned, it is the first time that a multi-objective derivative of the PSO algorithm is applied for optimal design of LQR controllers.

\section{Multi-Objective Quantum-Behaved PSO}

\subsection{Multi-Objective Optimization}
MOO is a common practice in engineering applications and refers to optimizing two or more competing, confronting or incommensurable objective functions simultaneously. In contrast to single-objective problems in which optimal solution is well-defined, in multi-objective problems it is not possible to reach the global optimum for all objectives at the same point. Instead there is a whole set of optimal trade-offs which forms the solution set called Pareto optimal set. Although it is possible to treat a problem with multi-objective nature as a single-objective problem, it is not an efficient approach due to two main reasons: (1) the objectives must be aggregated to a single weighted objective, and (2) the optimizer must run for several times to obtain the Pareto solutions. A MOO problem over an n-dimensional search space X is defined as finding the vector $X^*=\left[x_1^*,x_2^*,\ldots,x_n^*\right]\in X$ such that:  

\begin{equation}
\label{eq:6}
\left\{ \begin{array}{rcl}
\textnormal{Minimize} \quad\quad f_i(x),\quad\quad\quad  i=1\ldots k\\
\textnormal{Subject to} \quad g_j(x)\leqslant{0},\quad  j=1\ldots m
\end{array}\right.
\end{equation}

$f_i(x)$ and $g_j(x)$ are vectors of $k$ objective functions and $m$ constraints defined over $X$, respectively. Mostly it is not possible to find such a global and acceptable solution for all conflicting objectives. A feasible solution is to find a set of solutions in a way that each solution satisfies \ref{eq:6} without being dominated by other solutions. A solution $u = \left[u_1,\ldots,u_k\right]$ is said to dominate solution $v = \left[v_1,\ldots,v_k\right]$ if and only if $f_i(u) \leqslant f_i(v) \forall i = 1 \ldots k$ and $f_i(u) < f_i(v)$ for at least one objective function and is denoted by $u \prec v$. A feasible solution $x$ is Pareto optimal if and only if there is no other solution that dominates $x$. The set of all Pareto optimal solutions is referred as Pareto optimal set and the set of corresponding objective function values is called the Pareto front. The goal of a given MOO method is to find the Pareto optimal set which results in the best possible Pareto front regarding the given objectives and constraints.

\subsection{Particle Swarm Optimization}
The PSO algorithm \cite{47-eberhart1995new} is a population-based optimization technique that explores the search space by simulating the choreographed motion of the birds based on socio-cognitive characteristics of flocking individuals. A PSO algorithm consists of a swarm of $|S|$ particles each containing a position and a velocity vectors and a scalar fitness value. For a D-dimensional objective function, the position and velocity vectors of each particle are D-dimensional as well. The position vector represents the position of the particle within the D-dimensional search space and encodes a potential solution to the optimization problem at hand. The velocity vector determines the direction and magnitude of the changes in position vector through time. Each particle shares information with other particles through the best position seen so far in the swarm history and follows its trajectory toward the global optimum based on Newtonian mechanics. A domain-dependent fitness function is used to determine how strong a given particle is (i.e. how close the particle is to the global optimum). The velocity vector of the particle $i$ in a D-dimensional space is updated in through time using \ref{eq:7}.

\begin{equation}
\label{eq:7}
\vec{v}_i(t+1)=w(t) \times \vec{v}_i(t)+c_p \times \vec{r}_p \otimes \left( \vec{x}_{i,pbest}(t)-\vec{x}_{i}(t) \right) + c_g \times \vec{r}_g \otimes \left( \vec{x}_{gbest}(t)-\vec{x}_{i}(t) \right)
\end{equation}

$\vec{v}_i(t)$ denotes the D-dimensional velocity vector of the particle $i$ whose values are defined within the range of $[v_min,v_max]$. $w(t)$ is the inertia weight that controls the momentum of the particle by weighting the contribution of its previous velocity values. The inertia weight is equivalent to an infinite impulse response low pass filter. $c_p$ and $c_g$ are two positive parameters known as cognitive and social learning rates that determine the weight of cognitive aspects and social pressure in decision making process, respectively. $r_p$ and	$r_g$ are D-dimensional vectors of randomly generated values by a uniform distribution in the range of $[0,1]$. $\otimes$ denotes point-wise vector multiplication operator. $\vec{x}_{i,pbest}$ is the best position seen by the particle $i$ up to time $t$ ($pbest$),	
$xgbest(t)$ is the best position seen by whole swarm up to time $t$ ($gbest$), and	
$\vec{x}_{i}(t)$ is the position vector of particle $i$ in time $t$. The inertia factor is computed by \ref{eq:8}.

\begin{equation}
\label{eq:8}
w(t) = w_max - \left( w_max - w_min \right) \times t / T
\end{equation}

$w_max$ and $w_min$ are maximum and minimum values of inertia, respectively. $t$ and $T$ denote current iteration and maximum number of iterations, respectively. The position vector of particle $i$ is updated using \ref{eq:9}.

\begin{equation}
\label{eq:9}
\vec{x}_i(t+1)=\vec{x}_i(t)+\vec{v}_i(t+1)
\end{equation}

Without loss of generality, the best personal and the best global positions for a minimization problem are computed by \ref{eq:10} and \ref{eq:11}, respectively. In these equations, $f$ denotes the fitness function (i.e. performance index or cost function).

\begin{equation}
\label{eq:10}
\vec{x}_{i,pbest}(t+1)=\left\{ \begin{array}{rcl}
\vec{x}_{i,pbest}(t), \quad f\left(\vec{x}_{i}(t+1)\right)\geq f\left(\vec{x}_{i,pbest}(t) \right) \\
\vec{x}_{i}(t+1), \quad f\left(\vec{x}_{i}(t+1)\right) \leq f\left(\vec{x}_{i,pbest}(t) \right)
\end{array}\right.
\end{equation}

\begin{equation}
\label{eq:11}
\vec{x}_{gbest}(t+1)=\argmin_{\vec{x}_{i,pbest}} f\left( \vec{x}_{i,pbest}\left(t+1 \right)\right)
\end{equation}

The PSO algorithm has been exploited in optimizing various engineering problems such as electric power systems \cite{48-alrashidi2009survey}, wireless-sensor networks \cite{49-kulkarni2011particle}, image processing \cite{50-chander2011new,51-maitra2008hybrid}, robotics \cite{52-pugh2007inspiring,53-jatmiko2007pso}, to name only a few.

\subsection{Quantum-behaved PSO}
The main drawback of the PSO algorithm is that it does not guarantee the global convergence and is prone to the premature convergence \cite{54-van2006analysis}. The quantum-behaved PSO (QPSO)  \cite{55-sun2004particle,56-sun2004global,57-sun2005adaptive} is a global convergence guaranteed optimizer that belongs to the bare-bones PSO (BBPSO) category in which particles do not have a velocity vector and their position is updated by sampling a prob-ability distribution of interest. The QPSO was inspired by quantum mechanics and trajectory analysis of PSO \cite{30-sun2012quantum}. In \cite{58-clerc2002particle}, the trajectory analysis of the particles showed that each particle oscillates around its local attractor. It was also mathematically shown that if the upper limits of cognitive and social learning rates are selected properly, each particle will converge toward its local attractor. The local attractor is interpreted as the center of gravity toward which the particle careens while declining its kinetic energy. If the search space is stationary (i.e. which is the case in most of the practical applications) there will be no periodic orbits also known as unstable equilibria in the search hyperspace. Although satisfying this condition guarantees the convergence, there is no systematic way to select those limits. The local attractor of the $i$th particle denoted by $p_i$ is computed using \ref{eq:12}.

\begin{equation}
\label{eq:12}
\vec{p}_{i}(t)=\frac{\left(c_p\times\vec{x}_{i,pbest}(t) +c_g\times\vec{x}_{gbest}(t)\right)}{c_p+c_g}
\end{equation}

Contrary to PSO algorithm in which the particles follow their trajectory toward their local attractor based on Newtonian mechanics, in QPSO algorithm particles obey the quantum mechanics in which trajectory is not a valid concept. It is assumed that particles are attracted to a quantum potential field which is centered on their local attractors. In QPSO, the state of a particle is expressed by its wave function $\Psi\left(x,t\right)$. The probability of particle $i$ being in position $x_i$ is computed by probability density function $|\Psi\left(x,t\right)|^2$ whose form depends on the potential field in which the particle is moving. One possible model is quantum Delta potential well model \cite{55-sun2004particle}. In this model, it is assumed that a particle moves in a Delta potential well field in the search space with center $p$ calculated by \ref{eq:12}. In the quantum model, only the probability density function of the particle position is available while for computing the objective function the exact position of the particle is required. To address this issue,Monte Carlo sampling method is used to simulate the measure-ment process from the wave function which estimates the position of the $i$th particle as expressed by \ref{eq:13}.

\begin{equation}
\label{eq:13}
\vec{x}_{i}(t+1)=\vec{p}_{i}(t)\pm \frac{|\vec{x}_{i}(t)-\vec{p}_{i}(t)|\otimes ln\left( 1\setminus\vec{u}\right)}{g}
\end{equation}

In this equation, $p_i(t)$ denotes the local attractor of particle $i$ computed using \ref{eq:12}, $u$ is a D-dimensional randomly generated vector following a uniform distribution in the range of $[0,1]$, and $g$ is a control parameter greater than $ln\sqrt{2}$. The QPSO algorithm is summarized in Algorithm \ref{alg:1}.

\begin{algorithm}
\DontPrintSemicolon
\KwResult{returns the position vector of the global best particle}
\Begin{
Initialize the current positions randomly\;
\For{$t=1$ to $T$}{
	\For{$i=1$ to $|S|$}{
		Calculate fitness $f\left(\vec{x}_i(t)\right)$\;
		Update personal best $\left(\vec{x}_{i,pbest}(t)\right)$ using \ref{eq:10}\;
		Update global best $\left(\vec{x}_{gbest}(t)\right)$ using \ref{eq:11}\;
		$c_p,c_g \sim U[0,1]$\;
		Compute the local attractor $\vec{p}_i(t)$ using \ref{eq:12}\;
		\For{$d=1$ to $|D|$}{
			$u \sim U[0,1]$\;
			$L=\left(1/g\right)\times|\vec{x}_{i,d}(t)-\vec{p}_{i,d}(t)|$\;
			$\vec{x}_{i,d}(t)=\vec{p}_{i,d}(t)-L \times ln\left(1/u\right)$ with probability 	0.5\;
			otherwise $\vec{x}_{i,d}(t)=\vec{p}_{i,d}(t)+L \times ln\left(1/u\right)$\;			
		}
	}
}
\Return{$\vec{x}_{gbest}$}
} 
\caption{The standard QPSO algorithm.\label{alg:1}}
\end{algorithm}

$|T|$ and $|S|$ denote the number of iterations and swarm size,respectively. $U[0,1]$ is a uniform random number in the range of $[0,1]$ and $f(x)$ is an arbitrary objective function. The QPSO algorithm has some advantages over other variations of the PSO algorithm. It is less sensitive to the bias problem as it has only one parameter to be tuned. It is theoretically guaranteed that the QPSO converges to the global optimum. It is also experimentally shown that the QPSO algorithm outperforms other PSO variations in finding the optimal solutions and has stronger exploring capabilities \cite{30-sun2012quantum}. For more comprehensive evaluations on the QPSO algorithm, please refer to \cite{30-sun2012quantum}. The QPSO algorithm has been successfully applied to a vast variety of engineering problems such as but not limited to system identification \cite{59-fei2008parameters}, image processing \cite{60-lei2008two}, power systems \cite{61-zhisheng2010quantum}, neural network training \cite{62-li2007new}, brain-computer interfacing \cite{67-hassani2014incremental}, etc.

\section{Proposed Approach}
\subsection{Individual Representation}
In order to encode the problem of deciding the optimal weights of a given LQR, each particle is considered as a potential solution that encodes the elements of weighting matrices into its position vector. The state and control weighting matrices consist of $n^2$ and $m^2$ elements respectively which results in the total number of $n^2+m^2$ elements to be decided. Finding the optimum solution within a $(n^2+m^2)$-dimensional space is a challenging task due to the curse-of-dimensionality effect (e.g. 250 features for a system with 15 state and 5 control variables). A practical solution which is widely exploited in the real world applications is to define both matrices in diagonal form to reduce the search space to $(n + m)$ dimensions. Hence, as shown in \ref{eq:14}, the particles' position vector is defined as an $(n + m)$ dimensional vector that represents the concatenation of diagonal state and control weighting matrices.

\begin{equation}
\label{eq:14}
\vec{x}_{i}(t)=\left[x_{i,1}(t),\ldots,x_{i,n+m}(t)\right]=\left[Q_{1,1}(t),\ldots,Q_{n,n}(t),R_{1,1}(t),\ldots,R_{m,m}(t)\right]
\end{equation}

This simple trick reduces the number of features from $n^2+m^2$ to $n+m$ and as a result enhances the exploration capabilities and reduces the computational costs while reserving the accuracy.

\subsection{Evaluation Criteria}
To decide the evaluation criteria, the optimization goals are categorized to hard constraints and soft objectives. The former category refers to those constraints that must be satisfied. If a solution does not satisfy a given hard constrain, it is considered as an infeasible solution. On the other hand, those solutions that satisfy the hard constraints are feasible solutions that compete within the feasible sub-space to reach the better regions by optimizing the soft objectives. According to the definition in Section 2, Q and R matrices must be symmetric positive matrices and therefore those solutions with negative values are considered as infeasible solutions. This is the only hard constraint addressed in the proposed approach. The infeasible solutions are treated using a stochastic approach inspired by local search strategy in MA \cite{63-ishibuchi2003balance}. These solutions are either repaired with probability $P_r$ or debilitated by being penalized with probability $(1−P_r)$. The repair probability is defined by \ref{eq:15}.

\begin{equation}
\label{eq:15}
P_r\left(\vec{x}_{i}(t)\right)=1-\left(\sum_{j=1}^{n+m}\frac{H\left(x_{i,j}(t)\right)}{n+m}\right)
\end{equation}

In this equation, $x_i(t)$ denotes the position vector of particle $i$ in time $t$, $n$ and $m$ represent the cardinality of diagonal $Q$ and $R$ matrices respectively and $H(.)$ is the Heaviside step function that maps negative values to zero and positive values to one. The repair probability decreases proportional to the number of negative values within the position vector of a given particle. The repair strategy is thoroughly discussed in Section 5.3.

Three classes of soft objectives are considered as follows. (1) A given LQR controller must minimize the quadratic performance index $J$ defined by \ref{eq:3}. (2) An optimal controller also should minimize the time-domain control objectives that correspond to the transient response. The transient response of a system may be described using two competing factors including the swiftness and the closeness of the response to the desired response. The swiftness of the response is measured by the rise-time $Tr$. The closenessis measured by the overshoot $OS$ and settling-time $Ts$. (3) The steady-state behavior of the system is addressed by minimizing the steady-state error $Ess$. Furthermore, in order to augment the multi-objective aspect of the evaluation criteria, each particle is rewarded using\ref{eq:16} which is the difference between the number of solutions that are dominated by that particle and the number of solutions that dominate the particle.

\begin{equation}
\label{eq:16}
R\left(\vec{x}_i\right)=\left(\sum_{\forall x \in S,x \neq x_i} x_i \prec x \right) - \left(\sum_{\forall x \in S,x \neq x_i} x_i \succ x \right)
\end{equation}

The domination reward is normalized using a sigmoid function and aggregated with the violation penalty into a uniform weighted penalty-reward factor defined by \ref{eq:17}.

\begin{equation}
\label{eq:17}
f_{P-R}\left(\vec{x}_i\right)=\left(\frac{1}{1+e^{-|S|\times R\left(\vec{x}_i\right)}}\right)\times P_r\left(\vec{x}_i(t)\right)\times \phi
\end{equation}

In this equation, $f_{P-R}$ denotes the penalty-reward factor and $\phi$ denotes the penalty for violating the hard constraint. In order to aggregate the soft objectives in a multi-objective fashion, dynamic weighted aggregation (DWA) method in exploited in which the gradual changes (set by the change frequency $F$) of the weights force the particles to keep moving on the Pareto front. For a problem with two competing objectives, the dynamics of the weights is defined by \ref{eq:18}.

\begin{equation}
\label{eq:18}
w_1(t)=|\sin\left(2\pi t/ F \right)|, \quad w_2(t)=1-w_1(t)
\end{equation}

It has been shown that DWA outperforms bang-bang weighted aggregation (BWA) in case of convex Pareto front and is identical to BWA with concave Pareto front. DWA is also competitive with population-based non-Pareto approaches such as VEGA and VEPSO in terms of optimality while outperforms them in terms of efficiency \cite{64-parsopoulos2002particle}. The aggregated multi-objective criterion based on DWA is defined as shown in \ref{eq:19}.

\begin{equation}
\label{eq:19}
f\left(\vec{x}_i\right)=f_{P-R}\left(\vec{x}_i\right) \times \left(\left(w_1(t)\times\left(\log_{10}(J)+E_{ss}\right)\right)+\left(\left(w_2(t)\times\left(OS+T_s-T_R\right)\right)\right)\right)
\end{equation}

As shown in \ref{eq:19}, the quadratic performance index is normalized using a logarithmic operation. This is because the quadratic performance index is mostly much bigger than other elements in real world applications. The dynamically aggregated objective function shown in \ref{eq:19} lets the designer to decide the priority between minimizing the control effort and steady-state response, and optimizing the transient response while it automatically satisfies the hard constraint of the system. In other words, it provides the designer with a set of Pareto optimal solutions and lets her to decide the target solution based on practical preferences.

\subsection{Optimization process}
The standard QPSO algorithm introduced in \cite{55-sun2004particle} is augmented with two mechanisms including informed initialization and repairing strategy. Mostly, population-based heuristics construct an initial population of random individuals and then exploit an iterative optimization to push the population toward the global optima.The informed initialization refers to initializing the population based on a priori knowledge and has been shown that is more efficient than naive and random initialization \cite{65-dasgupta2008comparison}. In order to reinforce the algorithm with informed initialization, simulated annealing search \cite{66-ingber1993simulated} is exploited which is a fast search strategy that can escape the local optima due to its stochastic nature. This algorithm optimizes the problem at hand by emulating the cooling process of a solid until it reaches the minimum energy configuration. In this algorithm, in each step a random neighbor of the current state is chosen. In case that the selected neighbor is fitter than the cur-rent state, the search continues exploring from the neighbor state.Otherwise, it moves to the neighbor state with a probability proportional to the current temperature and the difference between the fitness values of the current and the neighbor states. This prob-ability decreases as the temperature reduces through the time. The proposed informed initialization based on simulated annealing is shown in Algorithm \ref{alg:2}.

\begin{algorithm}
\DontPrintSemicolon
\KwResult{returns a diverse and semi-optimal set of solutions}
\Begin{
Initialize population:$\vec{x}_i \sim U[0,X_{max}], \quad i=1\ldots|S|$\;
	\For{$i=1$ to $|S|$}{
		$w_1,w_2 \sim U[0,1]$\;
		\For{$t=1$ to $T_ini$}{
			\For{$d=1$ to $D$}{
				$Succ_d\left(\vec{x}_i\right)=\min\left(\max\left(N\left(\vec{x}_{i,d},\sigma\right),0\right),X_{max}\right)$ with probability $p_{Succ}$\;
			}
			Calculate $f\left(\vec{x}_i\right)$ and $Succ\left(\vec{x}_i\right)$ using \ref{eq:14} with current $w_1$ and $w_2$\;
			$\Delta E=f\left(\vec{x}_i\right)-f\left(Succ\left(\vec{x}_i\right)\right)$\;
			$T=e^{-\alpha t}$\;
			\If{$\Delta E>0$}{
				$f \left( \vec{x}_i \right) \leftarrow f \left( Succ \left( \vec{x}_i \right) \right)$
				}				
			\Else{
			$f\left(\vec{x}_i\right)\leftarrow f\left(Succ\left(\vec{x}_i\right)\right)$ with probability $e^{\Delta E/T}$\;
			}
		}
	}
\Return{$\vec{x}_{i}, \quad i=1\ldots|S| $}
} 
\caption{The informed initialization strategy.\label{alg:2}}
\end{algorithm}

The neighbor state $(Succ(x_i))$ is selected using Gaussian mutation operator adopted from evolutionary computation in which $N(x,\sigma)$ is a normal distribution of random variable $x$ with standard deviation of $\sigma$. The simulated annealing search is performed for $T_ini$ iterations on each particle which is fairly smaller than the number of iterations within the main optimizer. This process results in an initial swarm that is close to the Pareto optimal. It is noteworthy that to provide the initial swarm with a good degree of diversity,the aggregation weights are set randomly for each particle rather than using \ref{eq:18}.

The repair strategy is introduced to map the infeasible solutions into the feasible sub-space. As mentioned in Section 5.2, a particle with negative values within its position vectors violates the described hard constraint and is considered as an infeasible solution. A naive repair strategy is to randomly replace the negative values with random positive values. A more profound strategy is to exploit the strong feasible solutions to guide the infeasible particles
to those regions of feasible sub-space that are more likely to be close to the Pareto optimal. For this purpose, the cross-over and tournament selection operations are adopted from evolutionary computations. The repair strategy is as follows. For each infeasible particle, two feasible particles that are strongest in a randomly selected neighborhood are selected. Then, for each negative value embedded within the infeasible particles, the corresponding values of two selected particles are combined in a linear weighted manner. The negative value of the particle is then overwritten by the resulted value. The algorithm of repair strategy is shown in Algorithm \ref{alg:3}.

\begin{algorithm}
\DontPrintSemicolon
\KwData{$\vec{x}_i:$ position of infeasible particle $i$}
\KwResult{returns a feasible solution}
\Begin{
$\vec{p}_1,\vec{p}_2\leftarrow$ Tournament-Selection()\;
	\For{$d=1$ to $D$}{
		$\lambda \sim U[0,1]$\;
		\If{$\vec{x}_{i,d}<0$}{
		$\vec{x}_{i,d}=\lambda \vec{p}_{1,d}+\left(1-\lambda\right)\vec{p}_{2,d}$ with probability 0.5\;
		otherwise $\vec{x}_{i,d}=\left(1-\lambda\right) \vec{p}_{1,d}+\lambda\vec{p}_{2,d}$\;
			}
		}
\Return{$\vec{x}_i$}
} 
\caption{The repair strategy to map the infeasible particles to the feasible sub-space.\label{alg:3}}
\end{algorithm}

Using informed initialization and repair strategy, we introduce our proposed reinforced MO-QPSO algorithm for optimal tuning of LQR controllers in Algorithm \ref{alg:4}.

\begin{algorithm}
\DontPrintSemicolon
\KwResult{returns set of optimal diagonal Q and R matrices}
\Begin{
Informed-Initialization(): Algorithm \ref{alg:2}\;
	\For{$t=1$ to $|T|$}{
		\For{$i=1$ to $|S|$}{
			Calculate $P$ from Riccati equation: \ref{eq:5}\;
			Calculate the feedback gain $K$: \ref{eq:4}\;
			Simulate the closed-loop system: \ref{eq:2}\;
			Calculate fitness $f\left(\vec{x}_i(t)\right)$: \ref{eq:19}\;
			$Repair\left(\vec{x}_i(t)\right)$ with probability $P_r\left(\vec{x}_i(t)\right):$  \ref{eq:15} and Algorithm \ref{alg:3}\;
			Update personal best $\vec{x}_{i,pbest}(t)$: \ref{eq:10}\;
			Update global best $\vec{x}_{gbest}(t)$: \ref{eq:11}\;
			$c_p,c_g \sim U[0,1]$\;
			Compute the local attractor using $\vec{p}_i(t)$: \ref{eq:12}\;
			\For{$i=1$ to $|S|$}{
				$u \sim U[0,1]$\;
				$L=\left(1/g\right)\times|\vec{x}_{i,d}(t)-\vec{p}_{i,d}(t)|$\;
				$\vec{x}_{i,d}(t)=\vec{p}_{i,d}(t)-L \times ln\left(1/u\right)$ with probability 	0.5\;
				otherwise $\vec{x}_{i,d}(t)=\vec{p}_{i,d}(t)+L \times ln\left(1/u\right)$\;	
			}
		}
	}
\Return{all non-dominant particles}
} 
\caption{Proposed reinforced RMO-QPSO algorithm for optimal tuning of LQR controllers.\label{alg:4}}
\end{algorithm}

\begin{figure}
  \centering
  \includegraphics[scale=1.25]{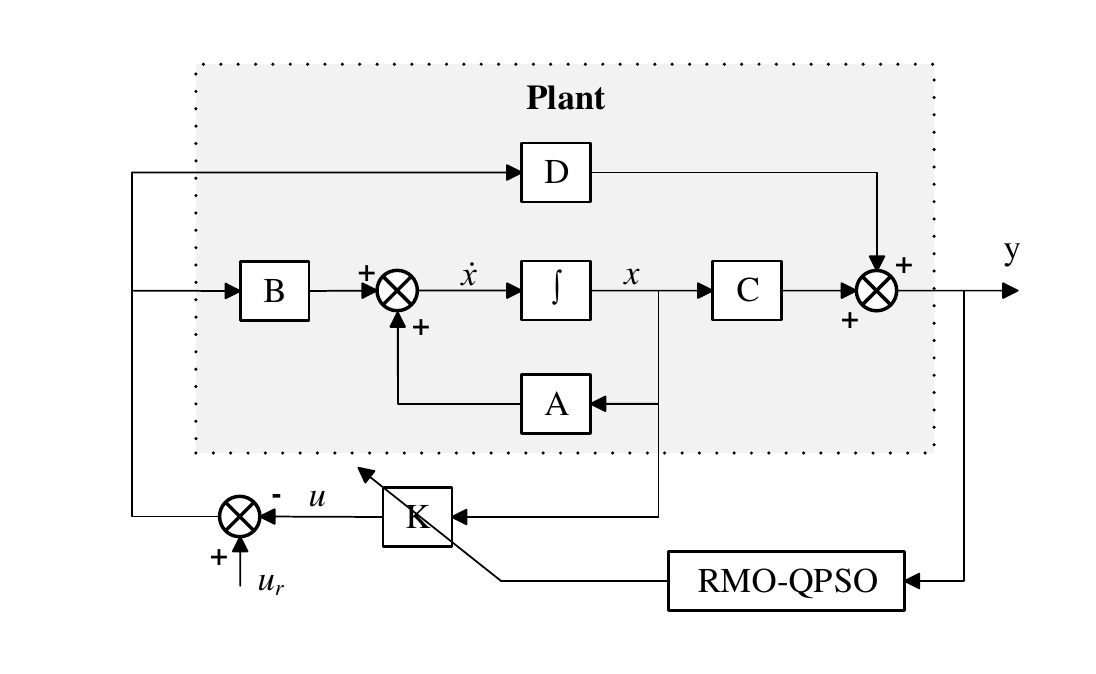}
  \caption{Schematic of the proposed RMO-QPSO within the closed-loop system.}
  \label{fig:Figure 1}
\end{figure}

In this algorithm, $|T|$, $|S|$, and $D$ denote the number of iterations, swarm size, and the number of diagonal elements of the weighting matrices, respectively. As shown, the proposed algorithm returns all non-dominant particles which build the Pareto optimal set. These Pareto optimal solutions are then presented to the designer and she decides the final solution based on some domain-dependent and practical considerations and her intuitions and expertise. The schematic of the integration of proposed RMO-QPSO algorithm into the closed-loop system is illustrated in Figure \ref{fig:Figure 1}.

\noindent

\section{Experimental results}
\subsection{Modeling}
\begin{figure}
  \centering
  \hspace*{-0.8cm}	\includegraphics[scale=0.75]{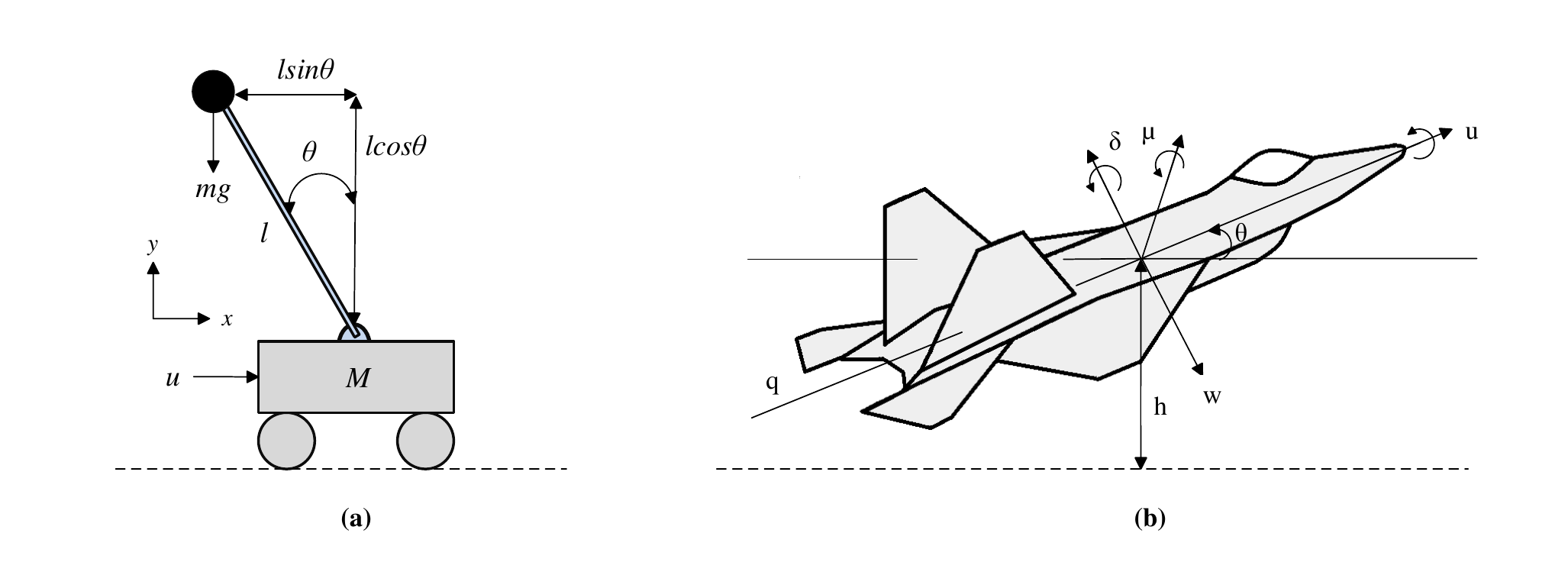}
  \caption{Schematics of evaluation benchmarks. (a) Inverted pendulum system with two degrees of freedom. (b) Aircraft landing flare system.}
  \label{fig:Figure 2}
\end{figure}

Without loss of generality, two classic control benchmarks are used for evaluating the introduced method. These benchmarks are an inverted pendulum system with two degrees of freedom and an aircraft landing flare system. The schematic of these two systems is illustrated in Figure \ref{fig:Figure 2}.

\noindent

The inverted pendulum system consists of a pendulum mounted on a movable cart which is restricted to linear motion. The control objective is to maintain the unstable equilibrium position by moving the cart along a horizontal track. The pendulum initially starts in an upright position with two acting forces including a vertical gravity force (mg) and an external horizontal force $u$. The state space representation of the system derived from Lagrange equation and linearized using Taylor series expansion is depicted in \ref{eq:20}.

\begin{equation}
\label{eq:20}
\begin{bmatrix}
\dot{x} \\ 
\dot{\upsilon} \\ 
\dot{\theta} \\ 
\dot{\omega}
\end{bmatrix}
=
\begin{bmatrix}
    0 & 1 & 0 & 0 \\
    0 & 0 & \frac{-mg}{M} & 0 \\
    0 & 0 & 0 & 1 \\
    0 & 0 & \frac{g\left(m+M\right)}{lM} & 0 \\
\end{bmatrix}
\times
\begin{bmatrix}
x \\ 
\upsilon \\ 
\theta \\ 
\omega
\end{bmatrix}
+
\begin{bmatrix}
0 \\ 
\frac{1}{M}\\ 
0 \\ 
\frac{-1}{lM}
\end{bmatrix}
\times
\begin{bmatrix}
    u
\end{bmatrix}
\end{equation}

In this equation, $x$ denotes the position of the cart and $v$ is its linear velocity, $\theta$ is the angle of the pendulum with respect to the vertical direction, and $\omega$ is its angular velocity. The simulation parameters are set as follows. The mass of the cart and pendulum are set to 0.5 kg and 0.2 kg, respectively. The length of the pendulum and the gravitational acceleration are set to 0.6 m and 9.81 $ms^−2$, respectively. The initial condition of the system is defined as $\left[x(0) \upsilon(0) \theta(0) \omega(0)\right]^T= [0 0 0 9]^T$. The second system is an aircraft landing flare system with the simplified motion shown in \ref{eq:21}.

\begin{equation}
\label{eq:21}
\begin{bmatrix}
\dot{u} \\ 
\dot{w} \\ 
\dot{q} \\ 
\dot{\theta} \\ 
\dot{h} \\
\dot{e}
\end{bmatrix}
=
\begin{bmatrix}
    −0.058 & 0.065 & 0 & −0.171 & 0 & 1 \\
    −0.303 & −0.685 & 1.109 & 0 & 0 & 0 \\
    0.072 & −0.685 & 0.947 & 0 & 0 & 0 \\
    0 & 0 & 1 & 0 & 0 & 0 \\
    0 & -1 & 0 & 1.133 & 0 & 0 \\
    0 & 0 & 0 & 0 & 0 & −0.571 \\
\end{bmatrix}
\times
\begin{bmatrix}
u \\ 
w \\ 
q \\ 
\theta \\ 
h \\
e
\end{bmatrix}
+
\begin{bmatrix}
    0 & 0 & −0.119 \\
   −0.054 & 0 & 0.074 \\
   −1.117 & 0 & 0.115 \\
    0 & 0 & 0 \\
    0 & 0 & 0 \\
    0 & 0.571 & 0 \\
\end{bmatrix}
\times
\begin{bmatrix}
   \mu \\
   \gamma \\
   \delta \\
\end{bmatrix}
\end{equation}

In this equation, $u$ denotes the longitudinal velocity, $w$ presents the vertical velocity, $q$ is the angular velocity about pitch with respect to the ground, $\theta$ denotes the pitch with respect to the ground, $h$ is the height, $e$ is the forward acceleration caused by throttle action, $\mu$ is the elevator angle, $\delta$ is the spoiler angle, and $\gamma$ is the throttle value. The simulated landing trajectory is described by $h'+ 0.2h = 0$ and the initial condition is defined as $[u(0) w(0) q(0) \theta(0) h(0) e(0)]^T= [5 −2.5 −1 −3 15 0.5]^T$. Considering the dynamics of this system, we use the integrated absolute error(IAE) measure rather than steady-state error. Hence, in \ref{eq:19} we replace the Ess term with IAE measure. Using landing trajectory and considering $h'= −w + 1.133 \times \theta$ from \ref{eq:21}, the IAE measure is computed using \ref{eq:22}.

\begin{equation}
\label{eq:22}
IAE=\int_0^{t_f}|e(t)|dt=\int_0^{t_f}|−w(t) + 1.133 \times \theta + 0.2 \times h(t)|dt
\end{equation}

\subsection{Experimental Setup with Sensitivity Analysis}
In order to conduct comparative studies on the performance of the proposed algorithm, the two benchmark systems are stabilized using the trail-and-error, LM, GA, DE, ABC, PSO, CPSO, AIWPSO, and RMO-QPSO methods. In order to eliminate the effect of bias (i.e.sensitivity of the meta-heuristics to their parameter setting) in the comparisons, sensitivity analysis based on full factor design \cite{31-lee2013sensitivity} is utilized. Although full factorial design is not feasible for systems with high degrees of freedom such as NN but it is completely tractable for most of the meta-heuristics as there are a few parameters to be decided. In order to carry out this task, each system is stabilized using a fixed set of parameter values for five times and the performance criterion is averaged. This process is repeated for all possible parameter settings and as soon as all the settings are evaluated, the setting that has resulted in the best performance is selected as the parameter setting for the meta-heuristic at hand. In order to generate the all possible parameter settings, a systematic and incremental approach is utilized. For each given parameter, a predetermined range is defined and then starting from the lower boundary, the value of the parameter is incremented by a defined step size until it reaches the upper boundary. These ranges and step sizes are defined as follows.

The range of iterations $T$ for all competing methods except trail-and-error is set to [25,150] with step size of 25. For LM optimizer, the sensitivity analysis is performed on the blending factor $\lambda$ within the range of [5,15] and step size of 1. For population-based meta-heuristics, the range of population size $|S|$ is set to[10,100] with increment factor of 10. For the GA, single point uniform cross-over, tournament selection mechanism with the neighborhood radius of 5 chromosomes, and Gaussian mutation operators are utilized. The cross-over probability $pc$ is defined in the range of [0.3, 0.9] with step size of 0.1 and the mutation probability $pm$ is defined in the range of [0.05, 0.3] with the step size of 0.05. Considering that contrary to GA, DE mostly relies on the mutation operator, the parameter setting for DE is define by switching the ranges of the cross-over and mutation probabilities used in GA. The ABC setting is as follows. The percentage of onlooker bees is set to 50\% of the colony, the employed bees are set to 50\% of the colony and the number of scout bees is selected as one. In the PSO algorithm, the range of the both cognitive and social parameters is set to [0.5, 2.0] with a step size of 0.1. The CPSO and AIWPSO settings are identical to the PSO setting. The chaotic sequence ofCPSO is generated by logistic maps. Finally, the control parameter of RMO-QPSO $g$ is set to the range of [0,1]. As suggested in literature \cite{55-sun2004particle}, the control parameter is updated by coefficient of $ln\sqrt{2}$.The results of parameter settings for different optimizers using full factorial sensitivity analysis are summarized in Table \ref{tab:1}. As shown, the best parameter settings for two benchmarks are selected using full factorial design. These acquired parameter settings will mitigate the bias sensitivity of the opponent methods and will result in fair comparisons.

\noindent
\begin{table}
\caption{Acquired parameter settings for simulations by sensitivity analysis.\label{tab:1}}
\noindent
\centering
\hspace*{-0.5cm}\begin{tabular}{|l|l|}
\hline
Method & Parameter setting  \\ 
\hline
Trail-error & --- \\
LM & $\lambda\leftarrow10, \quad T\leftarrow150$  \\
GA & $|S|=70, \quad T\leftarrow150, \quad p_c\leftarrow0.9,\quad p_m\leftarrow0.1$ \\
DE & $|S|=50, \quad T\leftarrow150, \quad p_c\leftarrow0.15,\quad p_m\leftarrow0.8$ \\
ABC & $|S|=30, \quad T\leftarrow175$ \\
PSO & $|S|=40, \quad T\leftarrow50, \quad c_p\leftarrow0.7,\quad c_g\leftarrow1.5$ \\
CPSO & $|S|=20, \quad T\leftarrow175$ \\
AIWPSO & $|S|=40, \quad T\leftarrow70, \quad w_{min}\leftarrow0.05,\quad w_{max}\leftarrow0.95$ \\
MO-QPSO & $|S|=20, \quad T\leftarrow75, \quad g\leftarrow 1.5ln\sqrt{2}$ \\
\hline
\end{tabular}
\end{table}

\subsection{Simulation Results}
The nine competitive methods are applied to stabilize the two benchmarks based on the parameter settings decided by sensitivity analysis. The trial-and-error method and the MA optimizer are run for one time as they do not have stochastic nature. On the other hand, the stochastic meta-heuristics are run for fifty times to approximate their behavior based on the acquired averages and standard deviations. The results are shown in Table \ref{tab:2}.

As shown in Table \ref{tab:2}, in case of inverted pendulum system,results indicate that all of the methods have successfully eliminated the steady-state error. In terms of overshoot the trial-and-error out-performs other methods. This is due to the slow response of the designed controller by expert. In terms of the rise-time RMO-QPSO and CSPO outperform other methods where as in terms of settling-time RMO-QPSO reaches the steady state faster than others. Also,RMO-QPSO and DE reach the best quadratic performance index.In controlling the flight landing system, RMO-QPSO outperforms other techniques in terms of integrated absolute error, overshoot,rise-time and settling time. In terms of minimizing the quadratic index CPSO and RMO-QPSO perform best.

The results suggest in the inverted pendulum system, RMO-QPSO enhances the rising time, settling time, and the quadratic performance index by 4\%, 8\%, and 2\%, respectively in comparison to the second best optimizer. Considering the fairly fast dynamics of the inverted pendulum system, these enhancements can boost the stabilization process in turn. Also, the results suggest that for the flight landing system, RMO-QPSO enhances the integrated absolute error, over-shoot, rising time, settling time, and the quadratic performance index by 8\%, 50\%, 70\%, 50\%, and 6\%, respectively. It is noteworthy that due to slower dynamics of the flight landing system, the proposed algorithm has resulted in greater enhancements in comparison with the inverted pendulum system. All in all, the results suggest that RMO-QPSO outperforms trail-and-error,gradient-based and stochastic techniques. It is also shown that ABC,DE, CPSO and AIWPSO achieve better performance in comparison to trial-and-error, ML, GA and PSO.

\noindent
\begin{table}
\caption{Acquired results from stabilizing two benchmark problems (inverted pendulum on the left and flight landing on the right).\label{tab:2}}
\centering
\noindent
\hspace*{-0.5cm}\begin{tabular}{|l|l||l|l|l|l|l||l|l|l|l|l|}
\hline
Optimizer& & $E_{ss}$ & $OS$ & $T_r$ & $T_s$ & $J$ & $IAE$ & $OS$ & $T_r$ & $T_s$ & $J$ \\
\hline
Trial-error & Mean & 0 & 0.23 & 0.46 & 2.09 & 26,000 & 31.45 & 0.22 & 9.25 & 18.23 & 24,500  \\
 &SD & –-- & –-- & –-- & –-- & –-- & –-- & –-- & –-- & –-- & –-- \\
LM & Mean & 0 & 0.51 & 0.49 & 1.84 & 24,200 & 26.08 & 0.21 & 9.03 & 14.32 & 26,700 \\
 &SD & –-- & –-- & –-- & –-- & –-- & –-- & –-- & –-- & –-- & –-- \\
GA & Mean & 0 & 0.85 & 0.54 & 1.19 & 22,000 & 20.32 & 0.19 & 6.53 & 12.31 & 24,200 \\
 &SD & 0 & 0.06 & 0.07 & 0.12 & 740.9 & 4.58 & 0.07 & 1.46 & 2.34 & 512.5 \\
DE & Mean & 0 & 0.62 & 0.46 & 1.11 & 20,500 & 14.35 & 0.11 & 2.51 & 11.22 & 24,220 \\
 &SD & 0 & 0.04 & 0.04 & 0.08 & 230.5 & 5.26 & 0.04 & 0.84 & 0.98 & 251.3 \\
ABC & Mean & 0 & 0.43 & 0.47 & 1.13 & 20,570 & 10.78 & 0.17 & 4.76 & 11.84 & 20,550 \\
 &SD & 0 & 0.02 & 0.06 & 0.08 & 268.0 & 1.56 & 0.04 & 0.79 & 1.87 & 98.5 \\
PSO & Mean & 0 & 0.89 & 0.48 & 1.11 & 21,600 & 11.21 & 0.18 & 5.23 & 11.97 & 22,000 \\
 &SD & 0 & 0.07 & 0.09 & 0.06 & 231.6 & 3.35 & 0.06 & 0.62 & 1.36 & 112.0 \\
CPSO & Mean & 0 & 0.46 & 0.43 & 1.15 & 20,700 & 10.06 & 0.19 & 4.89 & 11.93 & 19,300 \\
 &SD & 0 & 0.02 & 0.02 & 0.07 & 212.3 & 3.24 & 0.11 & 0.97 & 0.76 & 104.3 \\
AIWPSO & Mean & 0 & 0.59 & 0.47 & 1.20 & 20,850 & 9.64 & 0.15 & 3.91 & 11.46 & 20,580 \\
 &SD & 0 & 0.07 & 0.05 & 0.07 & 225.0 & 2.01 & 0.05 & 1.12 & 1.12 & 115.6 \\
RMO-QPSO & Mean & 0 & 0.67 & 0.43 & 1.02 & 20,500 & 8.89 & 0.07 & 1.42 & 7.27 & 19,300 \\
 &$\pm$ & 0 & 0.10 & 0.09 & 0.70 & 220.1 & 6.21 & 0.23 & 1.03 & 2.32 & 472.1 \\
\hline
\end{tabular}
\end{table}

In order to prove that the results shown in Table \ref{tab:2} carry statistical significance, pairwise one-tailed t-test is performed between the results of the RMO-QPSO and the other stochastic methods. The results of the t-test are summarized in Table \ref{tab:3}. In a standard t-test, the null hypothesis is rejected if the corresponding p-value is less than 0.05. As shown in Table \ref{tab:3},the t-tests performed among the results of different objective of RMO-QPSO and opponent stochastic methods reveals that all p-values are smaller than the threshold value. In other words, all the performed t-tests reject the null hypothesis and hence the simulation results are in fact statistically valid. 

\noindent
\begin{table}
\caption{Acquired results from stabilizing two benchmark problems (inverted pendulum on the left and flight landing on the right).\label{tab:3}}
\centering
\noindent
\hspace*{-0.5cm}\begin{tabular}{|l|l||l|l|l|l||l|l|l|l|l|}
\hline
Optimizer& & $OS$ & $T_r$ & $T_s$ & $J$ & $IAE$ & $OS$ & $T_r$ & $T_s$ & $J$ \\
\hline
GA&  p-value&  0.02&  0.01&  0.00&  0.00&  0.00&  0.01&  0.00&  0.02&  0.01 \\
DE&  p-value&  0.01&  0.00&  0.00&  0.00&  0.03&  0.00&  0.02&  0.00&  0.01 \\
ABC&  p-value&  0.00&  0.02& 0.04& 0.03& 0.00 &0.03 &0.02& 0.04 &0.00 \\
PSO&  p-value&  0.00& 0.02& 0.00& 0.00&  0.02&  0.04&  0.00& 0.03& 0.00 \\
CPSO&  p-value&  0.00& 0.00& 0.04& 0.01& 0.00& 0.03& 0.01& 0.00& 0.04 \\
AIWPSO&  p-value&  0.03& 0.04& 0.00& 0.00& 0.02& 0.00& 0.00& 0.01& 0.00 \\

\hline
\end{tabular}
\end{table}

\begin{figure}
  \centering
  \hspace*{-2.8cm}	\includegraphics[scale=0.75]{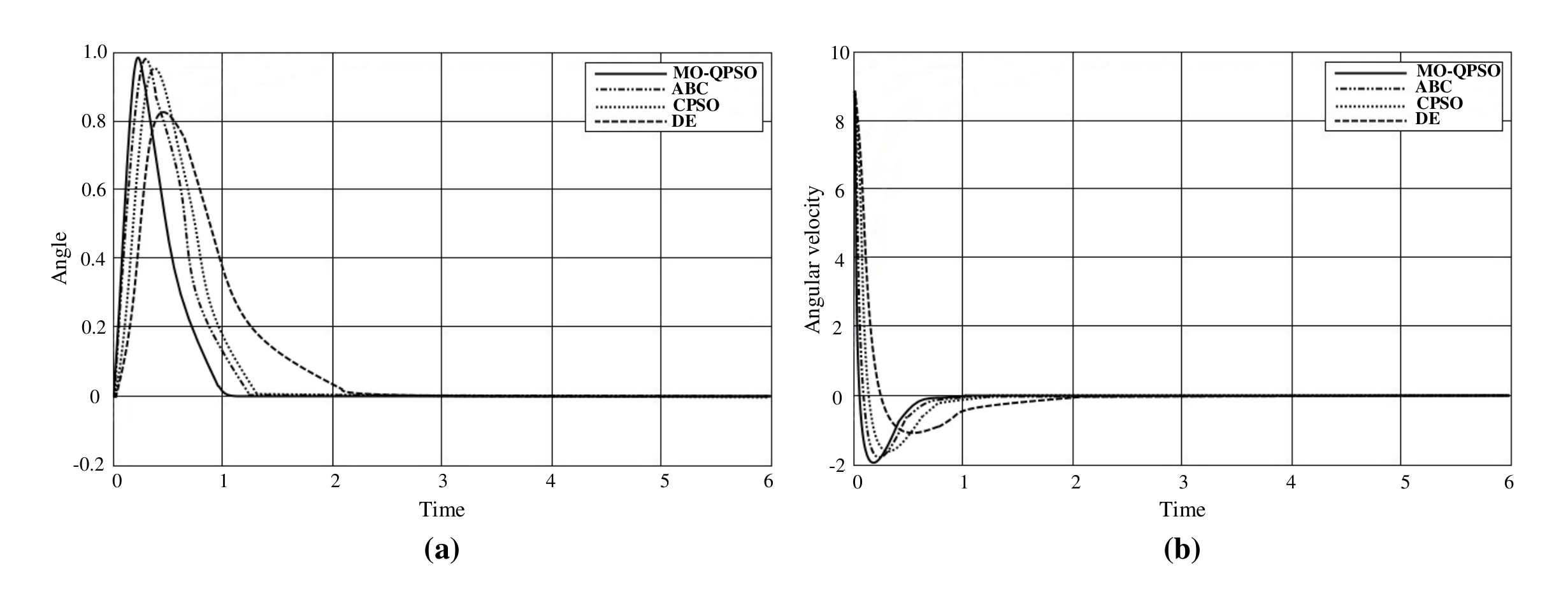}
  \caption{The responses of inverted pendulum system to the initial conditions. (a) Response of the angle. (b) Response of the angular velocity.}
  \label{fig:Figure 3}
\end{figure}

\begin{figure}
  \centering
  \hspace*{-2.8cm}	\includegraphics[scale=0.75]{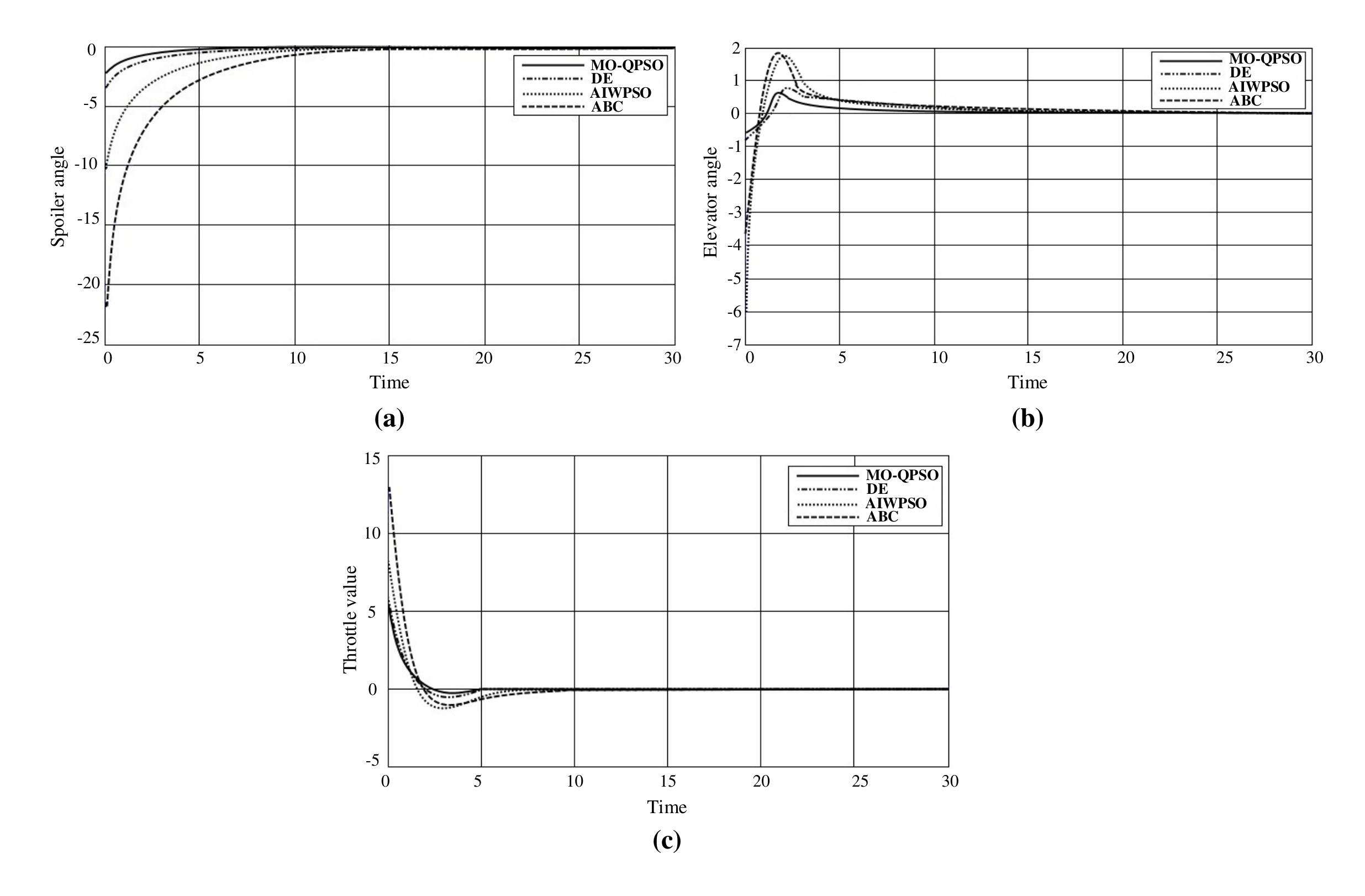}
  \caption{The responses of the flight landing system to the initial conditions. (a) Response of the spoiler angle. (b) Response of the elevator angle. (c) Response of the throttlevalue.}
  \label{fig:Figure 4}
\end{figure}

The responses of the pendulum's angle and its angular velocity to the initial conditions are depicted in Figure \ref{fig:Figure 3}. The responses of the spoiler angle, elevator angle, and throttle value of the flight landing system to the initial conditions are illustrated in Figure \ref{fig:Figure 4}. It is noteworthy that for the purpose of readability, the diagrams only illustrate the responses of the systems under the control of four best performed methods.

\section{Conclusion}
In this paper, a novel and generic multi-objective design paradigm is proposed which utilizes a global convergence guaranteed variation of particle swarm optimization (PSO) called quantum-behaved PSO (QPSO) for deciding the optimal configuration of the LQR controller for a given problem considering a set of competing objectives. There are three main contributions introduced in this paper as follows. (1) The standard QPSO algorithm is reinforced with an informed initialization scheme based on the simulated annealing algorithm and Gaussian neighborhood selection mechanism. (2) It is also augmented with a local search strategy which integrates the advantages of memetic algorithm into conventional QPSO. (3) Finally, an aggregated dynamic weighting criterion is introduced that dynamically combines the soft and hard constraints with control objectives to provide the designer with a set of Pareto optimal solutions and lets her to decide the target solution based on practical preferences. As far as the authors' knowledge is concerned, this is the first time that a multi-objective derivation of PSO is applied for deciding the configuration of LQR controllers.

The proposed method is compared against a gradient-based method, seven meta-heuristics, and the trial-and-error method on two control benchmarks including inverted pendulum system and flight landing system using sensitivity analysis and full factorial parameter selection and the results are validated using one-tailed T-test. The experimental results suggest that in inverted pendulum system, RMO-QPSO enhances the rising time, settling time, and the quadratic performance index by 4\%, 8\%, and 2\%, respectively in comparison to the second best optimizer. Also, the results suggest that for the flight landing system, RMO-QPSO enhances the integrated absolute error, over-shoot, rising time, settling time, and the quadratic performance index by 8\%, 50\%, 70\%, 50\%, and 6\%, respectively. All in all, the results suggest that RMO-QPSO outperforms trail-and-error, gradient-based and stochastic techniques. It is also shown that ABC, DE, CPSO and AIWPSO achieve better performance in comparison to trial-and-error, ML, GA and PSO. A possible direction for future studies is to enhance the RMO-QPSO in a way that it can address the online optimization of the LQR based on the signal stream.

\noindent

\noindent

\bibliographystyle{plain}
\bibliography{MOQPSO_bib}
\end{document}